\documentclass[runningheads]{llncs}

\usepackage{graphicx}
\usepackage{svg}
\usepackage{tabularx}
\newcolumntype{L}{>{\raggedright\arraybackslash}X}
\usepackage{array}
\usepackage{multirow}
\usepackage{subfig}
\usepackage{ulem}
\usepackage{caption}
\usepackage{hyperref}

\begin{document}
\title{Prior-RadGraphFormer: A Prior-Knowledge-Enhanced Transformer for Generating Radiology Graphs from X-Rays}
\titlerunning{Prior-RadGraphFormer}
\author{Yiheng Xiong\inst{1}$^*$ \and
Jingsong Liu\inst{1}$^*$ \and Kamilia Zaripova\inst{1}$^*$ \and Sahand Sharifzadeh\inst{2} \and Matthias Keicher\inst{1}\textsuperscript{\textdagger}
\and Nassir Navab\inst{1}\textsuperscript{\textdagger}
}
\def\thefootnote{*}\footnotetext{These authors contribute equally to this work and share first authorship.}
\def\thefootnote{\textdagger}\footnotetext{These authors share last authorship.}
\authorrunning{Xiong, Liu and Zaripova et al.}
\institute{Computer Aided Medical Procedures, Technische Universität München, Germany \and Ludwig Maximilians Universität München, Germany}
\maketitle              %
\begin{abstract}
The extraction of structured clinical information from free-text radiology reports in the form of radiology graphs has been demonstrated to be a valuable approach for evaluating the clinical correctness of report-generation methods. However, the direct generation of radiology graphs from chest X-ray (CXR) images has not been attempted. To address this gap, we propose a novel approach called Prior-RadGraphFormer that utilizes a transformer model with prior knowledge in the form of a probabilistic knowledge graph (PKG) to generate radiology graphs directly from CXR images. The PKG models the statistical relationship between radiology entities, including anatomical structures and medical observations. This additional contextual information enhances the accuracy of entity and relation extraction. The generated radiology graphs can be applied to various downstream tasks, such as free-text or structured reports generation and multi-label classification of pathologies. Our approach represents a promising method for generating radiology graphs directly from CXR images, and has significant potential for improving medical image analysis and clinical decision-making.
\end{abstract}
\section{Introduction}
In recent years, deep learning (DL) methods have significantly improved the computer-assisted diagnosis of chest X-ray (CXR) images. DL methods have been used in multiple ways. For instance, Ma and Lv~\cite{Pneumonia} used a Swin transformer with fully-connected layers to classify CXR images as either normal or indicative of pneumonia. Cicero et al.~\cite{Cicero2017TrainingAV} modeled the diagnosis process as a multi-label classification problem and used GoogleNet to classify CXR images into six categories. In comparison to these classification methods, generated free-text radiology reports can provide more comprehensive information about the impression and findings~\cite{radgraph}. Shin et al.\cite{LearnToRead} pioneered CNN-RNN for automatic text generation. Wang et al.\cite{TieNet} introduced TieNet, generating reports and detecting thorax diseases. Hou et al.\cite{RATCHET} proposed RATCHET, using transformer and attention to generate reports from CXR images. Kaur et al.\cite{kaur2022radiobert} used contextual word representations for useful radiological reports. Cao et al.\cite{cao2023mmtn} proposed Multi-modal Memory Transformer Network for consistent medical reports. Existing approaches for automating medical reporting rely on generating free text, posing challenges for clinical evaluation~\cite{pino2021clinically,few_shot}. To tackle this, ImaGenome~\cite{wu2021chest} and RadGraph Benchmark\cite{radgraph} were introduced to extract structured clinical information from free-text radiology reports, represented as a radiology graph. Each node in a radiology graph corresponds a unique entity, such as an object with bounding boxes annotations or an attribute in ImaGenome or, an anatomical structure or an observation and its presence and uncertainty in RadGraph Benchmark. Although these graph representations of reports have been used to evaluate the clinical correctness of reports~\cite{yu2022evaluating}, generating radiology graphs directly from CXR images has not been explored. In contrast, diverse interactions between object pairs in natural images in the form of scene graph generation have been extensively investigated \cite{lu2016visual}. Li et al.~\cite{Li_2022_CVPR}, and Lu et al.~\cite{lu2021seq2seq} employed two-stage methods to propose dense relationships between predicted connected object pairs. In recent work, Shit et al.~\cite{relationformer} introduced Relationformer, a unified one-stage framework based on DETR \cite{detr} that facilitates the end-to-end generation of graphs from images. It is a state-of-the-art method for detecting and generating graphs from natural images. However, it requires bounding boxes for the detected objects (nodes of the graph) and is not directly applicable to radiology graphs since some nodes (entities) in radiology graphs do not have exact locations, such as ``left`` or ``clear``, and most datasets do not provide bounding boxes annotations.\looseness=-1

Therefore, we propose a detection-free method, Prior-RadGraphFormer, to generate radiology graphs directly from CXR images without requiring bounding boxes for each entity. The method incorporates prior knowledge in the form of probabilistic knowledge graphs (PKG) \cite{cls_by_attention} that model the statistical relationship between anatomies and pathological observations. Experimental results show that Prior-RadGraphFormer achieves competitive results in the radiology-graph-generation task. Moreover, the generated graphs can be used for multiple downstream tasks such as generating free-text reports based on predefined rules, cheXpert labels \cite{chexpert} classification, and populating templates for structured reporting.

In summary, our contributions are the following: 1) proposing a novel detection-free method that generates radiology graphs directly from CXR images; 2) enhancing this method by incorporating prior knowledge, leading to improved performance; 3) extensively evaluating our method using RadGraph metrics and the two downstream tasks of report generation and multi-label classification of pathologies.
\section{Method}
\begin{figure}[htbp]
    \centering
    \includegraphics[width=0.8\linewidth]{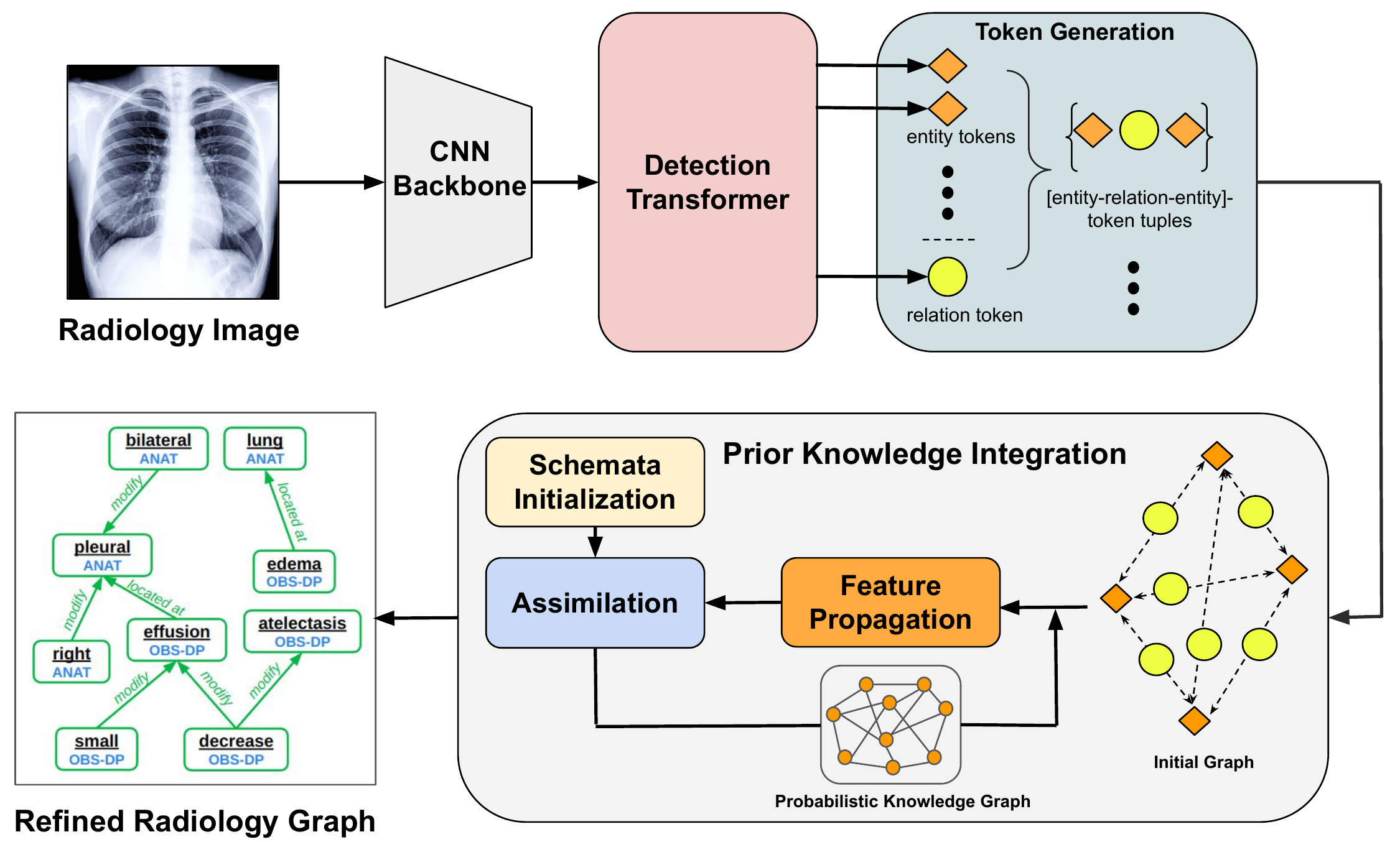}
    \caption{\textbf{Prior-RadGraphFormer architecture.} A CNN backbone extracts radiology image features, which are fed into the detection transformer (DETR)~\cite{detr} to generate multiple entity tokens and one relation token. Then an initial graph is constructed using entity and relation tokens. Each node is represented by a valid entity token and each edge is represented by a corresponding [entity-relation-entity]-token tuple. The initial graph is incorporated with prior knowledge and refined to output the final radiology graph. Our code is open sourced at \href{https://github.com/xiongyiheng/Prior-RadGraphFormer}{https://github.com/xiongyiheng/Prior-RadGraphFormer}}
    \label{model_architecture}
\end{figure}
We define radiology graph generation as the process of transforming CXR images into a graph structure introduced by RadGraph Benchmark~\cite{radgraph} that represents the content of a radiology report describing the image. Each node in the graph corresponds to a unique entity, such as an anatomical structure or an observation, and its presence and uncertainty. The edges between nodes indicate relationships between these entities. A visual depiction of the proposed method is shown in Fig.~\ref{model_architecture}. Prior-RadGraphFormer consists of two main components: RadGraphFormer and Prior Knowledge Integration.

\subsection{RadGraphFormer}
Shit et al.~\cite{relationformer} proposed Relationformer, an image-to-graph framework that leverages direct set-based object prediction and incorporates the interaction among the objects to learn an object-relation representation jointly. Given an input image, Relationformer initially outputs a set of discrete object tokens ([obj]-tokens) and a relation token ([rln]-token) where the token embeddings come from applying cross attention between a set of pre-defined and randomly initialized token vectors to input image patches. Relationformer then predicts the corresponding class label and the location of the bounding boxes for each object. In addition, Relationformer predicts a relation label for each pair of detected objects by concatenating each pair of [obj]-tokens with the [rln]-token and applying a relation prediction head on top.  

In our model, we adapt Relationformer as the radiology graph generation
backbone. We modify the entity prediction module by replacing the bounding box prediction head of it with the uncertainty prediction head due to the fact that radiology graphs do not contain bounding boxes but uncertainty information for each entity. Specifically, the entity prediction module comprises two distinct components. The first component is responsible for entity classification. The second component is responsible for predicting the uncertainty associated with each entity. Both are presented by one linear layer. To reflect these changes in terminology, we adjust the naming convention of the model's tokens, from [obj]-token to [ent]-token. Moreover, instead of using ResNet50 as the CNN backbone, we utilize
DenseNet121 ~\cite{densenet} based on its widespread adoption and effectiveness for processing
CXR images~\cite{RATCHET,few_shot}.

\subsection{Prior Knowledge Integration}
Vanilla RadGraphFormer uses the concatenation of a pair of [ent]-tokens and a shared [rln]-token $(\left\{ ent_{i},r,ent_{j} \right\}_{i \neq j})$, followed by an MLP, for relation prediction. Although the transformer-based model inherently considers the context of the tokens, we argue that these representations may not adequately capture the complexity of the relationships. Consequently, this limited representation may lead to an incomplete or inaccurate understanding of the context, resulting in sub-optimal relation and/or entity prediction.

Following~\cite{cls_by_attention}, in order to address this issue, we propagate higher-level prior knowledge in RadGraphFormer, as shown in the lower part of Fig.~\ref{model_architecture}. To this end, we first construct an initial embedding graph $G$ from all the valid $ent_{i}$s as nodes. Each edge in the graph is represented by $edg_{ij} = MLP_{proj}(\left\{ ent_{i},r,ent_{j} \right\}_{i \neq j})$, where $MLP_{proj}$ helps project the concatenated edge features to the same dimension of node features. This results in a fully-connected bi-directional graph. Then a stack of graph transformers $GTs$~\cite{graph_transformer} are utilized to propagate features of both nodes and edges in $G$. In order to allow the storage and propagation of relational knowledge within our framework, we create a randomly initialized representation for each class as a trainable parameter $s_c$ called schemata~\cite{cls_by_attention} that interacts with the outputs of the graph transformer. We then apply multiple assimilation steps~\cite{cls_by_attention}. The assimilation step is a process where the outputs of $GT$ attend to $s_c$s such that the attention coefficients predict the classification outputs for each node/edge from the $GT$ and the attention values are propagated to the output of $GT$. It is important to note that the attention coefficients are supervised by the ground truth labels for each entity/relation during the training. Details can be found in the appendix.\looseness=-1

\subsection{Training and Inference.} During training, we apply two supervisions for entity classes. One is for [ent]-tokens generated from DETR and the other is for the attention coefficients of nodes during each assimilation step. The latter one can be regarded as an additional entity class supervision. Entity uncertainty is supervised via [ent]-tokens and relation is supervised via the attention coefficients of edges during each assimilation step. During inference, we evaluate entity metrics solely based on [ent]-tokens. As for relation inference, we use the attention coefficients of edges from the last assimilation step as the classification output.
\section{Experiments}
\subsection{Datasets}\label{datasets}
Our dataset comprises MIMIC-CXR-JPG v2.0.0~\cite{physionet,mimic-cxr-jpg,mimic-cxr-jpg-2}, which contains both imaging studies and free-text reports, and RadGraph Benchmark~\cite{radgraph}. Our training set includes around 220,000 ground truth graph annotations obtained by RadGraph Benchmark, paired with the corresponding frontal CXR images. The validation set consists of 500 ground truth graph annotations obtained by board-certified radiologists and the corresponding frontal CXR images. There are 229 classes of entities after mapping, three levels of uncertainty, and three classes of relation. Details can be found in the appendix.

\subsection{Implementation Details}  All networks including the backbone are trained from scratch with PyTorch 1.12.0 and CUDA 11.6 on a single NVIDIA A40 until convergence. A batch size of 32 is chosen, and AdamW~\cite{adamw} optimizer with a learning rate of 1e-4 is utilized. To address the issue of imbalanced labels in entity class prediction, we employ focal loss~\cite{focal_loss}, while cross entropy loss~\cite{crossentropyloss} is used for entity uncertainty prediction. For relation prediction, we utilize stochastic relation loss~\cite{relationformer}, with a foreground-to-background edge ratio of 1:3. The weights between entity classification loss, uncertainty loss, and relation loss are 1, 1, and 3. Additionally, two assimilation steps are performed during the prior knowledge integration process per iteration. CXR images are preprocessed following~\cite{few_shot}.

\subsection{Evaluation} We evaluate our model similarly to RadGraph Benchmark~\cite{radgraph} considering micro F1-score. A predicted entity is regarded as a true positive if its predicted entity class and uncertainty level are correct. For relation evaluation, a predicted relation is considered a true positive if its head entity, tail entity, and relation type are predicted correctly. To show the use cases for radiology graph generation in this work, we perform the evaluation for the two downstream tasks: free-text report generation and cheXpert labels~\cite{chexpert} classification. Specifically, we generate free-text clinical reports from generated radiology graphs based on pre-defined rules.\looseness=-1
\begin{figure}[ht]
     \centering
    \includegraphics[width=0.9\linewidth]{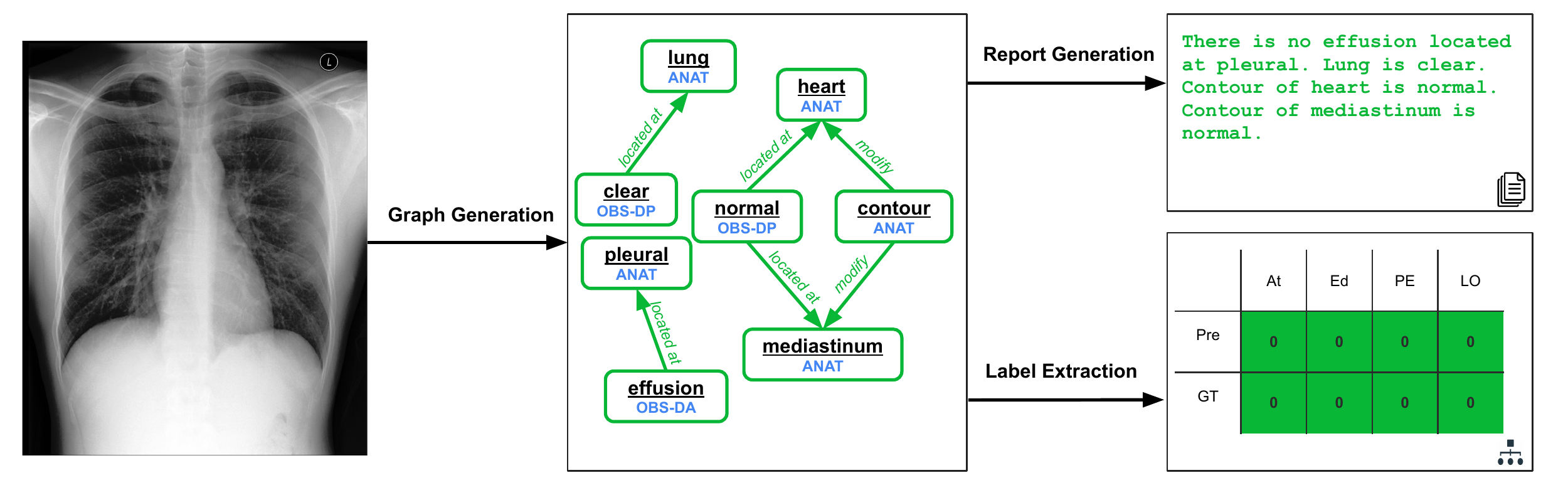}
     
    \includegraphics[width=0.9\linewidth,,page=2]{viz/figure2-2_croped.pdf}
  \caption{\textbf{Qualitative results}. On the left and middle are the CXR images and the corresponding predicted radiology graphs by our model. On the right, we present the outputs of two downstream tasks derived via the predicted graph: free-text clinical report and pathologies classification result. True positives, false negatives, and false positives are represented by green, gray, and red, respectively.}%
  \label{qualitative_results}
\end{figure}

\begin{table}[ht]
    \centering
    \caption{\textbf{Results comparisons on the validation set.} (P: precision, R: recall; B-1: BLEU-1, MET.: METEOR, R\_L: ROUGE\_L scores; At: Atelectasis, Ed: Edema, PE: Pleural Effusion, LO: Lung Opacity.) Best quality performance in bold.}
    \resizebox{\linewidth}{!}{
    \begin{tabular}{c|c|c|c|c|c|c|c|c|c|c|c|c|c|c}
    \hline
    \hline
    \multirow{3}*{Method} & \multicolumn{6}{c|}{RadGraph Metrics} & \multicolumn{4}{c|}{NLP Metrics} &  \multicolumn{4}{c}{Classification F1}\\
    \cline{2-15}
    & \multicolumn{3}{c|}{Entity} & \multicolumn{3}{c|}{Relation} & \multirow{2}*{B-1} & \multirow{2}*{MET.} & \multirow{2}*{R\_L} & \multirow{2}*{SPICE} & \multirow{2}*{At} & \multirow{2}*{Ed} & \multirow{2}*{PE} & \multirow{2}*{LO}\\
    \cline{2-7}
    & P & R & F1 & P & R & F1 & & & & & & & & \\
    \hline
    RATCHET & - & - & - & - & - & - & \textbf{0.18} & \textbf{0.09} & 0.19 & \textbf{0.09} & 0.38 & 0.27 & 0.47 & 0.17 \\
    RATCHET $\rightarrow$ RadGraph Benchmark & 0.45 & 0.16 & 0.24 & 0.11 & 0.03 & 0.05 & - & - & - & - & - & - & - & - \\
    vanilla RadGraphFormer & 0.58 & 0.47 & 0.52 & 0.35 & 0.19 & 0.25 & 0.14 & 0.08 &\textbf{0.21} & 0.08 & \textbf{0.40} & 0.26 & \textbf{0.82} & \textbf{0.24} \\
    Prior-RadGraphFormer & \textbf{0.63} & \textbf{0.55} & \textbf{0.59} & \textbf{0.44} & \textbf{0.21} & \textbf{0.28} & 0.07 & 0.07 & 0.19 & 0.08 & 0.39 & \textbf{0.29} & 0.81 & 0.22 \\
    \hline
    \hline
    \end{tabular}
    }
    \label{main_result}
\end{table}

\section{Results and Discussion} 

\subsection{Main Results}
Given that no prior work has addressed our task, we create a baseline that can be evaluated via RadGraph metrics. The baseline is constructed by two separate pretrained models, where RATCHET{~\cite{RATCHET}} is used to predict free-text reports from CXR images and followed by RadGraph Benchmark{~\cite{radgraph}} to produce radiology graphs from predicted reports. Note that radiology graphs generated from ground truth reports via pretrained RadGraph Benchmark are essentially ground truth labels as discussed in Section. {\ref{datasets}} hence they are not directly comparable with our results. Additionally, we compare the performance of vanilla RadGraphFormer with Prior-RadGraphFormer. As is shown in Table.~\ref{main_result}, both vanilla RadGraphFormer and Prior-RadGraphFormer demonstrate superior performance over the baseline in all RadGraph metrics. In particular, Prior-RadGraphFormer shows better performance than vanilla RadGraphFormer, indicating the significance of leveraging prior knowledge for radiology graphs generation. 

\subsection{Downstream Tasks}
For downstream tasks, we compare our methods with RATCHET~\cite{RATCHET} using free-text NLP metrics and cheXpert label classification F1 score. However, please note that there may be variations in the numerical results of RATCHET due to differences in the validation set used in our experiments compared to the original paper. Table.~\ref{main_result} shows that our radiology graphs directly generated from CXR images can be properly transformed into free-text reports and pathologies classification results. They have reasonable performance in the corresponding evaluation and demonstrate the utility of a graph-based representation in the context of a clinical study. 

From Table.~\ref{main_result} it is evident that vanilla RadGraphFormer exhibits better performance with respect to BLEU-1 score and F1 score of several pathologies as compared to Prior-RadGraphFormer. This disparity in performance can be explained by a more varied set of entities in radiology graphs generated by vanilla RadGraphFormer.
Also, it is worth noting that the NLP metrics of RATCHET are generally better than our methods. However, higher NLP metrics may not indicate better clinical usefulness~\cite{pino2021clinically}. One example is shown in Table.~\ref{compare_metric_study}. Qualitative results can be seen in Fig.~\ref{qualitative_results}. 

\begin{table}[ht]
    \centering
    \caption{\textbf{Example of evaluating clinical reports with NLP metrics.} In this case, our report exhibits superior clinical accuracy compared to RATCHET, but NLP metrics associated with it are significantly lower.}
    \begin{tabularx}{\linewidth}{L|c|c|c|c|c}
    \hline
    \hline
       \multirow{2}*{Report section} & \multicolumn{4}{c|}{NLP Metrics} & \multicolumn{1}{c}{Classification}\\
       \cline{2-6}
       &B-1 & MET. & R\_L & SPICE & PE\\
       \hline
       \textbf{Ground Truth}: As compared to the previous radiograph, small left and moderate  layering right pleural effusions have increased in size. & - & - & - & - & Positive\\
       \hline
       \textbf{RATCHET}: As compared to the previous radiograph, the patient has been intubated. & 0.36 & 0.19 & 0.47 & 0.24 & Negative \\
       \hline
       \textbf{Prior\&Vanilla-RadGraphFormer}: There is enlarged effusion located at bilateral of pleural. & 0.07 & 0.06 & 0.13 & 0.12 & Positive\\
    \hline
    \hline
    \end{tabularx}
    \label{compare_metric_study}
\end{table}

\begin{table}[ht]
    \centering
    \caption{\textbf{Ablation studies on the validation set.} (AECS: additional entity class supervision)}\looseness=-1
    \resizebox{\linewidth}{!}{
    \begin{tabular}{c|c|c|c|c|c|c}
    \hline
    \hline
       \multirow{2}*{Method} & \multicolumn{3}{c|}{Entity} & \multicolumn{3}{c}{Relation}\\
       \cline{2-7}
       & Precision & Recall & F1-score & Precision & Recall & F1-score\\
       \hline
       vanilla RadGraphFormer & 0.58 & 0.47 & 0.52 & 0.35 & 0.19 & 0.25\\
       Prior-RadGraphFormer w/o PKG & 0.61 & 0.50 & 0.55 & 0.39 & 0.17 & 0.23 \\
       Prior-RadGraphFormer w/o AECS & 0.63 & 0.49 & 0.55 & 0.43 & 0.17 & 0.24 \\
       Prior-RadGraphFormer & \textbf{0.63} & \textbf{0.55} & \textbf{0.59} & \textbf{0.44} & \textbf{0.21} & \textbf{0.28} \\
    \hline
    \hline
    \end{tabular}
    }
    \label{ablation_study}
\end{table}

\subsection{Ablation Studies} In our ablation studies, we aim to investigate two key aspects. Firstly, we aim to determine whether the integration of PKG truly improves performance or whether it is the graph transformers that enhance the model. Secondly, we aim to explore the impact of incorporating additional entity class supervision on entity and relation metrics. 

 To assess the impact of PKG integration, we simply use nodes features and edges features after graph transformers without assimilation to generate classification outputs. The results presented in Table.~\ref{ablation_study} demonstrate that Prior-RadGraphFormer without PKG is inferior to Prior-RadGraphFormer, thereby highlighting the importance of incorporating PKG as priors in Prior-RadGraph-Former.

 As for the influence of incorporating additional entity class supervision in the assimilation step, we conduct a comparative experiment by removing it and analyzing the resulting performance. The experimental results displayed in Table.~\ref{ablation_study} show that the impact is positive.

 \subsection{Limitations and Outlook} In this work, we evaluated the performance without directly using the graph itself. In the future work we plan to investigate additional graph-specific metrics \cite{rolinek2020deep,wills2020metrics}  to provide a more comprehensive evaluation. Furthermore, we acknowledge that the number of baselines is limited and aim to explore more comparative models in our ongoing research. Additionally, the inclusion of PKG increases the training time. In future work, we plan to optimize the integration of PKG to accelerate the training without compromising performance.
\section{Conclusion}
In this paper, we propose Prior-RadGraphFormer, a novel detection-free method that generates radiology graphs directly from CXR images. The model incorporates prior knowledge in the form of probabilistic knowledge graphs that capture the statistical relationship between anatomies and observations. We demonstrate the effectiveness of Prior-RadGraphFormer in the CXR-image-to-radiology-graph task. Moreover, we show that generated radiology graphs are useful for downstream tasks such as free-text reports generation and multi-label classification of pathologies. Our findings offer a promising direction for automatically generating radiology graphs from CXR images. We pave the way to automate the classification of fine-grained clinical findings structured in a radiology graph that can be applied to generate reports while allowing for the assessment of clinical correctness.

\subsubsection{Acknowledgements}
The authors gratefully acknowledge the financial support by the Federal Ministry of Education and Research of Germany (BMBF) under project DIVA (FKZ 13GW0469C). 
Kamilia Zaripova was partially supported by the Linde \& Munich Data Science Institute, Technical University of Munich Ph.D. Fellowship.
\bibliographystyle{splncs04}
\bibliography{main.bbl}
\appendix
\vspace{1cm}
\section*{Supplementary material}
\begin{table}[htbp]
    \centering
    \caption{\textbf{Sample entity mappings.}}
    \resizebox{\linewidth}{!}{
    \begin{tabular}{c|c}
        \hline
        \hline
        clear & clear, lucencies, lucency \\
        \hline
        consolidation & consolidation, consolidations, consolidative \\
        \hline
        congested & congested, congestion, engorged \\
        \hline
        sharp & sharp, sharply \\
        \hline
        prominent & prominent, prominence \\
        \hline
        nodule & nodule, nodules, nodular \\
        \hline
        pneumonic & pneumonic, pneumonia \\
        \hline
        calcification & calcification, calcified, calcifications \\
        \hline
        tortuous & tortuous, tortuosity \\
        \hline
        atelectasis & atelectasis, atelectatic, atelectases \\
        \hline
        \hline
    \end{tabular}
    }
    \label{entity_mapping}
\end{table}

\begin{figure}[htbp]
    \centering
    \includegraphics[width=1.0\linewidth]{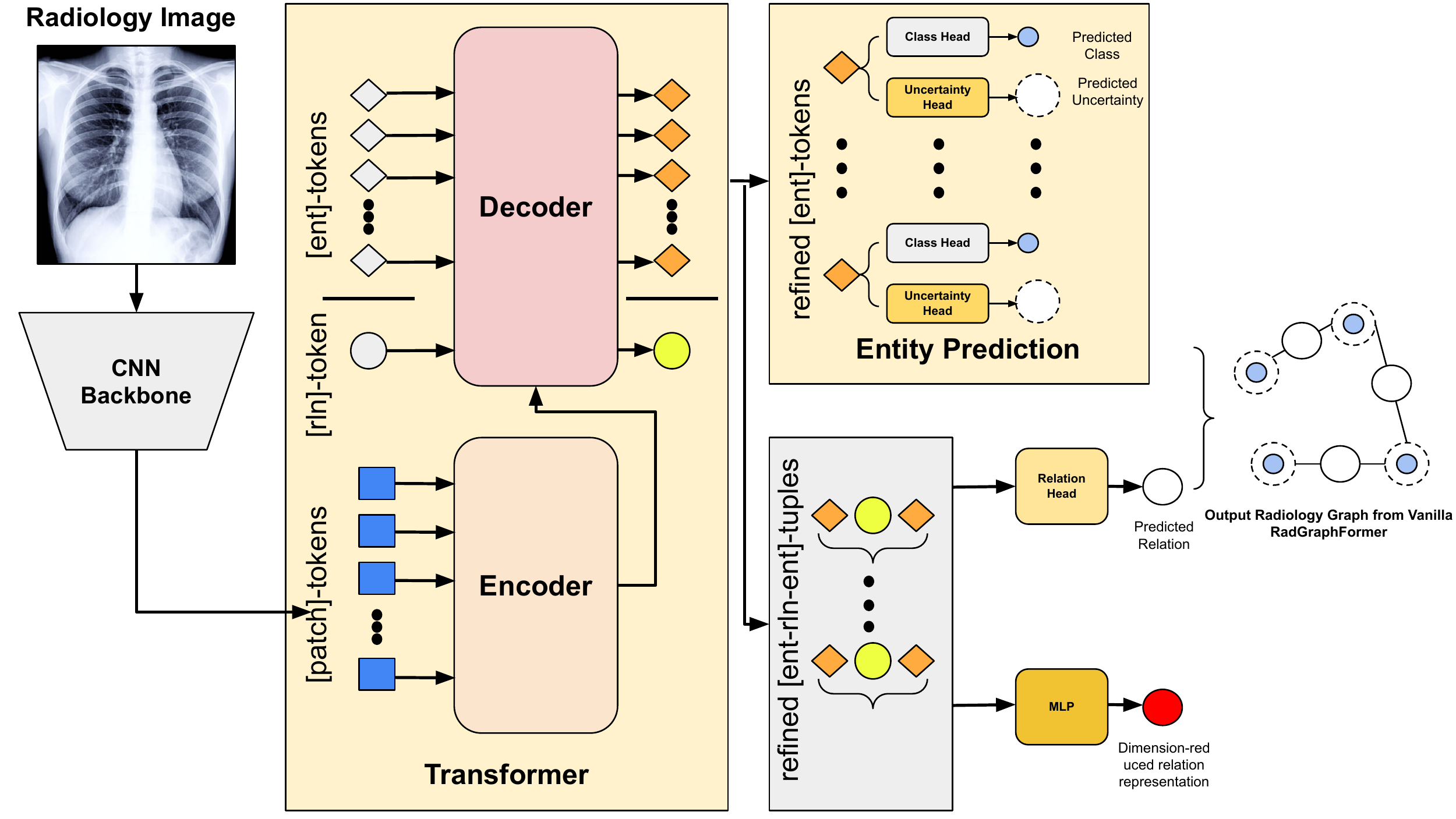}
    \caption{\textbf{Vanilla RadGraphFormer architecture.} Vanilla RadGraphFormer passes the concatenation of a pair of [ent]-tokens and a shared [rln]-token (concatenated edge feature) to a relation head to predict the relation type. For constructing the initial embedding graph for prior knowledge integration, an MLP is used to reduce the concatenated edge feature to make both nodes and edges dimension identical.}
    \label{vanilla}
\end{figure}

\begin{figure}[htbp]
    \centering
     \includegraphics[width=1.0\linewidth]{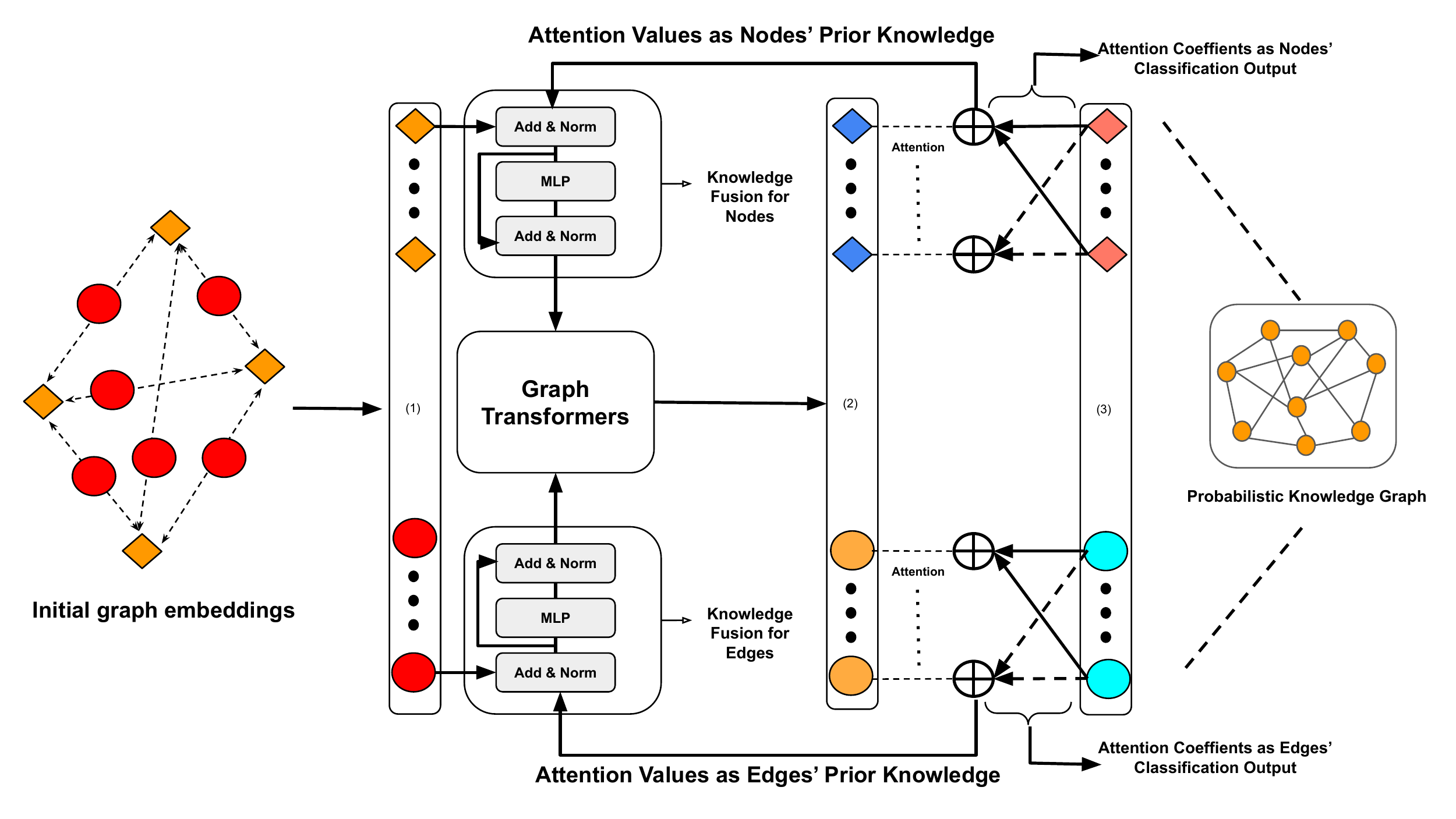}
    \caption{\textbf{Prior knowledge integration process.} (1) Initial representations from vanilla RadGraphFormer; (2) Propagated features by graph transformers; (3) Randomly initialized schema representations.}
    \label{PKI}
\end{figure}
\end{document}